\documentclass[letterpaper, 10 pt, conference]{ieeeconf}  

\IEEEoverridecommandlockouts                              

\usepackage{math}
\usepackage{amsmath}
\usepackage{booktabs}
\usepackage{cite}

\title{\LARGE \bf MTG: Mapless Trajectory Generator with Traversability Coverage for Outdoor Navigation}

\author{
Jing Liang$^{1}$, Peng Gao$^{1}$, Xuesu Xiao$^{2}$, Adarsh Jagan Sathyamoorthy$^{1}$, \\
Mohamed Elnoor$^{1}$, Ming C. Lin$^{1,3}$ and Dinesh Manocha$^{1}$
\thanks{$^{1}$University of Maryland at College Park. $^{2}$George Mason University. $^{3}$Amazon.}
}

\begin{document}

\maketitle
\thispagestyle{empty}
\pagestyle{empty}

\begin{abstract}
We present a novel learning-based trajectory generation algorithm for outdoor robot navigation. Our goal is to compute collision-free paths that also satisfy the environment-specific traversability constraints. Our approach is designed for global planning using limited onboard robot perception in mapless environments while ensuring comprehensive coverage of all traversable directions. Our formulation uses a Conditional Variational Autoencoder (CVAE) generative model that is enhanced with traversability constraints and an optimization formulation used for the coverage. We highlight the benefits of our approach over state-of-the-art trajectory generation approaches and demonstrate its performance in challenging and large outdoor environments, including around buildings, across intersections,  along trails, and off-road terrain, using a Clearpath Husky and a Boston Dynamics Spot robot. In practice, our approach results in a $6\%$ improvement in coverage of traversable areas and an $89\%$ reduction in trajectory portions residing in non-traversable regions. 
Our video is here: \url{https://youtu.be/3eJ2soAzXnU}
\end{abstract}

\section{INTRODUCTION}

Global planning for autonomous mobile robots has evolved significantly over the years. While many methods rely on map-based planning~\cite{ganesan2022global, gao2019global, psotka2023global},  accurate maps are not always available for many scenarios. These scenarios include rural areas without complex RGB or geometric features in the environment~\cite{ort2018autonomous}, areas under construction that undergo significant change over time, satellite-inaccessible areas, etc. In these cases, mapless global planning~\cite{ort2018autonomous} or real-time map analysis~\cite{schmid2020efficient, lluvia2021active} are used for navigation. 

\begin{figure}[t]
    \centering
    \includegraphics[width=0.95\linewidth]{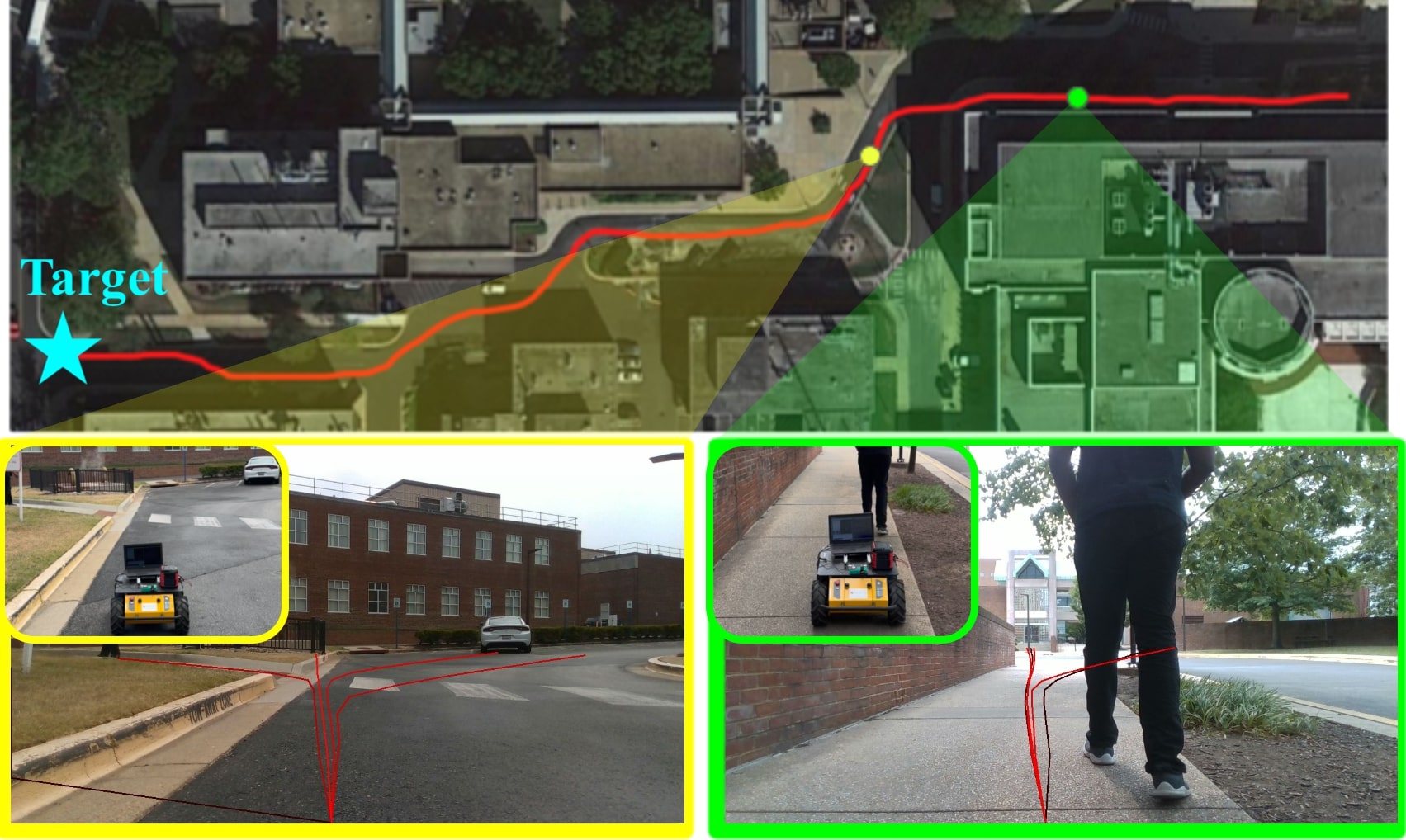}
    \caption{Trajectory generation in a campus environment. The top view is the robot's trajectory (around 350m). The bottom row shows generated trajectories at the two locations corresponding to green and yellow boxes, respectively. The trajectory closest to the global target, shown at the top, will be chosen to provide waypoints for the local planner. The local planner drives the robot and avoids collisions~\cite{liang2022adaptiveon}. In this complex outdoor scenario, our MTG global navigation method can efficiently and safely generate trajectories for global navigation and cover most of the traversable areas, including the narrow pedestrian sidewalks.
    }
    \label{fig:all_trajectory}
    \vspace{-1em}
\end{figure}

Most work in map-based planning has focused on computing a collision-free optimal trajectory~\cite{gasparetto2015path, ozturk2022review, guo2023efficient}.  In contrast, a key issue in mapless global planning is to compute traversable directions for long-range navigation and combine that with collision-avoidance capabilities of a local planner~\cite{ort2018autonomous, dobrevski2021deep}. These traversable directions can be calculated at a low frequency (0.1 Hz) to provide a high-level trajectory for robots to follow. The difference between outdoor map-based and mapless navigation arises from the fact that the lack of an accurate global map does not allow a robot to conduct very accurate collision checking {\em with occlusions}. Therefore, the global planner only needs to compute the rough navigation directions at different locations for outdoor navigation. This task can be rather difficult, as the robot needs to estimate the possible trajectories based on the limited onboard observations on complex, outdoor terrains.

The trajectory generation task for global planning presents several challenges, especially in the absence of a map. As shown in Figure \ref{fig:all_trajectory}, on one hand, the generated trajectories should avoid non-traversable areas, such as bushes, trees, etc. On the other hand, the generated trajectories should be able to cover all these traversable directions in the view of the robot, account for partial occlusion caused by static and dynamic obstacles, and resolve any ambiguities in path computation, as shown in Figure~\ref{fig:all_trajectory} (bottom right).

Current solutions for off-road global planning typically separate trajectory generation and environmental traversability analysis~\cite{fan2021step, Fankhauser2018ProbabilisticTerrainMapping, sathyamoorthy2022terrapn}. Moreover, they often rely on intuitive human maps to handle the occlusion issues and analyze the traversability of the environment. However, robots could potentially leverage encoded information, which can be generated by encoding the observations of the robot with some learning models, to directly generate trajectories without necessitating pre-conceptional map construction. Furthermore, map construction and planning are also computationally intensive~\cite{shan2020lio}, which hinders real-time navigation.  Therefore, our work aims to develop an end-to-end model that seamlessly handles occlusion, traversability analysis, and trajectory generation together. To this end, generators~\cite{cvae, gupta2018social} provide a solution for empirical trajectory generation. For example, DLOW~\cite{yuan2020dlow} and Ma et al.~\cite{ma2021likelihood} utilize historical trajectories to generate new ones, incorporating likelihood loss to encourage the diversity of the generated trajectories. However, they often overlook inter-trajectory influences and traversability constraints.

\noindent {\bf Main Results:} We present a novel approach for trajectory generation in mapless global planning, focusing on creating trajectories that cover the most viable directions within a robot's limited (120-degree) field of view while respecting environmental traversability constraints. We present an \textbf{end-to-end} mapless trajectory generator, MTG, which can efficiently compute viable trajectories that are traversable for both wheeled and legged robots. Our model is trained and demonstrated in the outdoor scenes, where the non-traversable areas include bushes, trees, buildings, streets with traffic, etc. The MTG approach achieves {\bf 72\%} coverage of the traversable areas in the robot's field of view. The coverage is calculated by the distance between generated trajectories and ground truth trajectories (see Section~\ref{sec:approach}). The major contributions include: 

\begin{enumerate}
\item \textbf{Novel End-to-End Learning Model:} We propose a novel end-to-end architecture for global trajectory generation, which can efficiently and effectively generate viable trajectories. Based on the architecture of the Conditional Variational Autoencoder (CVAE), we extend the method to the task of using multiple trajectory distributions to cover possible traversable areas. Different distributions of the embedded information, generated by the encoder, are applied with the attention mechanism to provide the conditions to the decoders. The attention model informs the decoders about other trajectories, guiding the generation of more effective trajectories, which are closer to the ground truth.
\item \textbf{New Loss Functions:} We design an innovative loss function that accounts for multiple constraints inherent in mapless trajectory generation. These include traversability, diversity, and coverage constraints. Our approach ensures trajectories account for the occlusions caused by dynamic obstacles and do not travel into non-traversable regions while performing comprehensive exploration of open outdoor areas.

\item  \textbf{Performance Improvement:} Empirical evaluations reveal that our model outperforms existing methods, such as DLOW~\cite{yuan2020dlow} and CVAE~\cite{cvae}. We demonstrate the performance by conducting tests on various robots, including  Clearpath Husky and Boston Dynamics Spot robots, in challenging environments with occlusions by dynamic obstacles and different non-traversable areas, such as trees, buildings, bushes, etc. The testing environment for global planning has a scale of several hundred meters. Our method consistently shows {\bf at least 6\%} improvement in coverage of traversable areas and up to {\bf 89\%} reduction in trajectory portions belonging to non-traversable regions.
\end{enumerate}

\section{BACKGROUND}
Path generation in traversable regions in outdoor navigation has attracted significant attention from the robotics community~\cite{canny1988complexity,manocha1992algebraic,lavalle2006planning} In this section, we review different approaches.

\textbf{Traversability Analysis:}
The simplest approach to path generation is to separate the task into traversability analysis and planning~\cite{fan2021step, gasparino2022wayfast, jonasfast}. For traversability analysis, multiple sensors can be used as input. Frey et al.~\cite{jonasfast} use an RGB camera and ViT~\cite{caron2021emerging} with self-supervised learning to segment traversable areas; WayFAST~\cite{gasparino2022wayfast} uses an RGBD sensor and a Resenet backbone to segment the traversable areas from the input observations. Step~\cite{fan2021step} uses both a Lidar and an RGB mono camera to extract an elevation map and analyze multiple risks on the map, including collisions and steep elevation. These methods assume that with the perfect traversability map, the trajectory generation problem can be handled by regular path/motion planning algorithms. However, in real-world scenarios, building an accurate traversability map in real-time is non-trivial due to occlusions in observations and the computational constraints in building a map~\cite{Fankhauser2018ProbabilisticTerrainMapping}. In addition, the map is for human perception, but a robot does not require a human-friendly map for navigation. We don't need to generate any intermediate map in the process. Therefore, an end-to-end learning strategy could be efficient to handle the problem of trajectory generation in traversable areas with occluding obstacles.

\textbf{End-to-end Approaches: } Generating trajectories on traversable areas can be based on the empirical information of either other agents or the robot itself. TridentNetV2~\cite{paz2022tridentnetv2} uses a raw path planned by a global planner as a guide to help the trajectory generation reach a target specified by a GPS coordinate. However, this method still requires a high-level map for planning before trajectory generation. ViKiNG~\cite{shah2022viking} and LaND~\cite{kahn2021land} train their models to generate trajectories approximating a collected dataset, but these methods are trained with a specific optimization target and cannot provide multiple possible directions for different navigation purposes.
FlowMap~\cite{flowmap} assesses the agents in front of the robot to estimate the traffic flow and to generate the trajectories. However, all these methods require some guidance beforehand and cannot fully explore the traversable areas by generating multiple trajectories. Our approach, on the other hand, generates diverse trajectories that satisfy traversability constraints and can cover most traversable areas in front of the robot.

\textbf{Trajectory predictions:} Trajectory prediction estimates the future trajectories of agents based on their historic paths. SocialGAN~\cite{gupta2018social} and Flomo~\cite{scholler2021flomo} use generative models to forecast the trajectories of the agent, but they cannot generate diverse trajectories to cover all traversable areas. DLOW~\cite{yuan2020dlow} and Ma et al.~\cite{ma2021likelihood} use the likelihood loss to learn diverse trajectories. However, they do not constrain the trajectories to traversable areas. For terrains with limited traversable areas, such methods could experience convergence issues since some trajectories always lie in non-traversable regions. Vern~\cite{sathyamoorthy2023vern}, TerraPN~\cite{sathyamoorthy2022terrapn}, GrASPE~\cite{weerakoon2022graspe} and AdaptiveON~\cite{liang2022adaptiveon} handle the navigation in off-road navigation scenarios. Although these methods only solve short-distance motion planning problems, they are complementary to our approach. 

\begin{figure*}
    \centering
\includegraphics[width=0.95\linewidth]{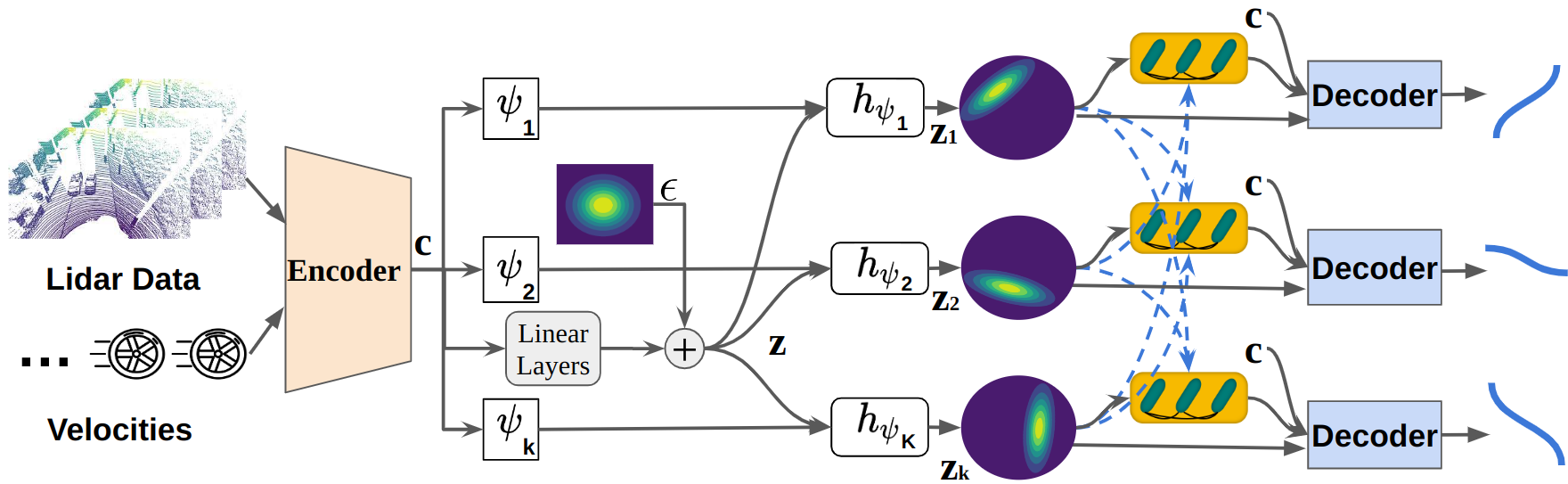}
    \caption{\textbf{Overall Pipeline of MTG:} The inputs are several consecutive frames of Lidar point clouds and velocities of the robot, $\epsilon$ is a Normal distribution, and $\c$ is the condition value. Green circles represent distribution of $\z_k$, $p(\z_k)$, where $k\in\set{1,...,K}$. The last part represents the decoders, and each decoder generates one trajectory.
    }
    \label{fig:pipeline}
    \vspace{-1em}
\end{figure*}
\section{APPROACH}
\label{sec:approach}
In this section, we formulate our problem and describe our approach to solving it. For global navigation, we need the local trajectory generator to provide enough trajectory candidates to cover the traversable areas in the available free space. In this method, we only use a map in the training stage. In the experiment stage, the method directly takes robots' perceptions as input and output trajectories. During training, with a traversability map shown in Appendix~\cite{appendix} Section \ref{sec:a_start_generated}, we use the A* algorithm to generate trajectories to different distributed targets, which are in traversable areas with a certain distance $D$ from the robot. We assume that these trajectories mostly cover the directions to all open spaces in the $D$-meters. We define these trajectories as the ground truth trajectories. The coverage of the generated trajectories is measured by the distances to the ground-truth trajectories.

\subsection{Problem Formulation}
Given the observation $o \in \co$, our end-to-end model, $m_\theta(\cdot)$ with parameters $\theta$, generates trajectories $\tau_k \in \ct$, where $\tau_k = m_\theta(o)_k$ is the $k_\text{th}$ output of the model $m_\theta(o)$. K is the number of generated trajectories. We formulate trajectory generation as an optimization problem under the constraints of traversability as follows:
\begin{align}
    \hat{\theta} = \argmax{\theta} {\left(\frac{\bigcup_{k=1}^K f(m_\theta(o)_k)}{\abs{\ca}} - \beta \sum_{k=1}^K g(m_\theta(o)_k, \Tilde{\ca})\right)} ,
    \label{eq:definition}
\end{align}

\noindent where $\hat{\theta}$ is final well-trained parameter of our model. $f(\cdot)$ represents the area covered by the generated trajectories $\tau_k$ in the traversable area, $\ca$. $g(\cdot)$ calculates the portion of the path $\tau_k$ in non-traversable area, $\Tilde{\ca}$. $\beta$ is a hyper-parameter. Intuitively, the first term is to encourage trajectories to cover traversable areas, and the second term is to constrain their traversability. In this approach, we use the negative exponential distance to the ground truth to substitute the first term of Equation~\ref{eq:definition}. 

In our formulation, the input contains two sorts of observations $\co=\set{\cl_i, \cv}$, where $\cl_i$ represents the past $N_l$ frames of 3D Lidar point clouds and $\cv$ composes $N_v$ consecutive frames of the robot's velocities. The encoder, as shown in Figure \ref{fig:pipeline}, processes the observation into an embedding vector $\z$ and a condition vector $\c$. Some randomness is introduced for the trajectory generation by Gaussian noise $\epsilon$ to $\z$. Then a decoder takes the embedding vector to generate a trajectory $\tau$. The model's formulation is represented as,
\begin{align}
    p(\tau|\c) &= \int_\z p(\tau|\z,\c) p(\z)d\z ; \\
    p(\tau|\c) &\approx \frac{1}{S}\sum_{s=1}^S p(\tau|\z^{(s)},\c), \;\;\;\z^{(s)} \sim p_\theta(\z|\x).
\end{align}
Here, $\tau \in \ct$ denotes the generated trajectory, and its distribution form is denoted as $ p(\tau | \x)$. $\tau$ is composed of S waypoints in the forward (X-axis) and the left (Y-axis) directions of the current robot frame. $\x\in \co$ denotes the input observation. In addition, $\l \in \cl$, $\v\in\cv$. $\z\in \cz$ denotes random sampling data from a Gaussian distribution $p(\z)$.
We use a network to represent the function $\c=f_e(\x)$. $f_e$ represents the encoder.

\subsection{Attention-based CVAE}
\label{sec:attention_approach}

From Equation~\ref{eq:definition}, there are two major tasks: 1. generate trajectories to cover all traversable areas; 2. make the trajectories traversable. For the 1st task, it requires the trajectories to be as diverse as possible to maximize the coverage. To make the trajectories diverse, we utilize the DLOW~\cite{yuan2020dlow} algorithm to linearly project the embedding vector $\z$ to different axes and train the model with trajectories' distance. For the 2nd task, considering we can generate many trajectories but the traversable areas might be limited, there could be some redundant trajectories that cannot find uncovered traversable areas. Thus, we define the trajectories closest to the ground truth trajectories as \textit{effective} trajectories and other trajectories as \textit{redundant} trajectories, under the assumption that ground truth trajectories are the smallest number of trajectories to cover all the traversable areas. In this project, the output of each decoder is a sequence of positional difference $(\Delta x_n, \Delta y_n)$, which can be accumulated to a trajectory $\tau = \set{p_1,...,p_n,...,p_N}$, where each position $p_n=(x_n, y_n)$. 

For the diversity of the trajectories, as DLOW, we use multiple invertible linear transformations $\z_k = A_i(c)\z + b_i(c)= h_{\psi_ari}(\z)$ to transform the embedding Gaussian distribution $\cn(\mu, \nu)$ to different Gaussian distributions $\cn(\mu_k, \nu_k)$, where $\psi_i = \set{A_i(\c), b_i(\c)}$ represents the parameters of the linear transformation layers. $\z$ is the embedding Guassian distribution with randomness $\epsilon \sim \cn(\zero, \I)$. For K trajectories, each trajectory $\tau_k$ can be generated by the decoder, $p(\tau_k|\z_k, \c)$, where $\z_k\in \cz$ and $\cz$ represents the set of Gaussian distributions.

For the effective and redundant trajectories, given the embeddings $\cz$, we use a self-attention model~\cite{vaswani2017attention} $g(\cz)$ to calculate the relationship among all $\z_k\in\cz$. Therefore, each embedding $\z_k$ is enhanced by other embeddings' information $\Bar{\cz}$, where $\Bar{\cz}_k = \cz \setminus \z_k$.  The enhanced embedding $\z_k$ will be put into a decoder to generate a single trajectory. Now we have the CVAE formulation as 
\begin{align}
    p(\tau_k | \c) &= \int_{\z_k} p \left ( \tau_k | \z_k, \c, \Bar{\cz}_k \right ) d\z_k.
\end{align}

This formulation is to constrain the effective trajectories $\tau_c$ and redundant trajectories $\tau_o$ to be as close as possible, and $\tau_{c, i} \in \tau_c$ are as diverse as possible to cover all ground truth trajectories. Therefore, here are two targets: $\min d(\tau_c, \tau_o)$ and $\max d(\tau_{c,i}, \tau_{c,j})$, where $d(\cdot,\cdot)$ is the distance function between two trajectories. DLOW~\cite{yuan2020dlow} solves the second target, but it's not able to solve the first target effectively due to the lack of information on other trajectories. Our formulation provides the information $\Bar{\cz}$ to the decoders and helps trajectories to achieve the first target.

\subsection{Constraints on Traversability, Diversity, and Coverage}
The generated trajectories should satisfy different constraints in the outdoor environment: 1. cover different traversable directions; 2. only lie in traversable areas; and if there are redundant trajectories after covering all the traversable areas, these trajectories should still lie in traversable areas. These constraints are not trivial challenges, so in the following, we present the methods to apply the constraints as loss functions in our approach.

\textbf{The CVAE's lower bound loss function:} This loss function is to constrain the generated trajectories as ground truth trajectories. The  lower bound loss function contains two parts: KL-divergence and reconstruction loss, as in Equation \ref{eq:cvae_loss}
\begin{align}
    \cl_{C} = -\cd_{KL} (q (\z|\x) \parallel p (\z)) + \frac{1}{L} \sum_{l=1}^{L} \log p_\theta(\tau|\x,\z).
    \label{eq:cvae_loss}
\end{align}
The reconstruction loss can be formulated in two parts: 1. The difference between the final positions from both the ground truth and the generated trajectory; 2. The average Hausdorff function~\cite{aydin2020usage}to make the generated trajectories $\hat{\tau}$ cover all the target trajectories ${\tau}$, as in Equation \ref{eq:hausdorff}, where $s$ and $\hat{s}$ are waypoints in the trajectories.
\begin{align}
    d_h(\hat{\tau}, \tau) = \frac{1}{2} \left ( \frac{1}{\abs{\tau}}\sum_{s\in \tau}  \min_{\hat{s}\in \hat{\tau}}d(s,\hat{s})+ \frac{1}{\abs{\hat{\tau}}}\sum_{\hat{s}\in \hat{\tau}}\min_{s\in \tau}d(\hat{s},s) \right ).
    \label{eq:hausdorff}
\end{align}

In contrast to the traditional CVAE reconstruction loss with only one ground truth and one output, we have multiple ground truth and output trajectories in each step. We want the generative model to cover all the ground truth instead of only one-one comparison. Therefore, we change the reconstruction loss to the coverage loss, where in each time step we only back-propagate the information of the nearest trajectory to each of the ground truth trajectories:
\begin{align}
    \cl_{CVAE} = \cd_{KL} (q (\z|\x) \parallel p (\z)) +  d_h(\hat{\tau}_c, \tau_k),
    \label{eq:fianl_cvae_loss}
\end{align}
where the $\hat{\tau}_c$ is the closest trajectory to the $k_{th}$ ground truth trajectory $\tau_k$.

\label{item:diversity} \textbf{Diversity loss function:} In this problem, we generate T trajectories at each step that are diverse enough. Very different from DLOW, if T is more than the ground truth trajectories, we want the redundant generated trajectories as close as possible to other ground truth trajectories. We have the diversity loss function:
\begin{align}
    \cl_{D} = \exp{(-d(\hat{\tau}_c, \hat{\tau}_c))} + \exp{(d(\hat{\tau}_o, \hat{\tau}_c))},
\end{align}
where $\hat{\tau}$ is the output trajectory, $\tau_c$ represents the trajectories closest to one of the ground truth trajectories, and $\tau_o$ represents the redundant trajectories. The redundant $\tau_o$ always appears when generated trajectories are more than ground truth trajectories. $d(\cdot)$ is the Euclidean distance between two trajectories, where each $\tau_o$ only compares with the closest $\tau_c$. The attention model contributes to telling the information of other trajectories and enhances the generated trajectories either close to the ground truth or close to other effective trajectories, $\tau_c$.

\textbf{Traversability loss function:} Trajectory generation requires the generated trajectories to be collision-free with nearby obstacles (bushes, trees, buildings, etc) or incursions into non-traversable areas. We use the distance to the non-traversable areas as the loss to train the generated trajectories away from the regions:
\begin{align}
    \cl_c = \exp{\left(1 - \min(\max(\frac{1}{N}\sum_{n=1}^N\min{d(\cc, \p_n)},1),0)\right)},
\end{align}
where $\cc$ represents the obstacles set near the robot and distance function $d(\cdot)$ represents the Euclidean distance between the generated waypoint $\p_n$ and the obstacles or non-traversable areas.

The total loss function can be written as,
\begin{equation}
\begin{aligned}
    \cl = \beta_1 \cd_{KL} (q (\z|\x) \parallel p (\z)) + \beta_2 \left( d_h(\hat{\tau}_c, \tau_k) +
    \cl_D \right) + \beta_3 \cl_c,
\end{aligned}
\end{equation}
where the first part is the CVAE KL-divergence, and the second part contributes to the exploration by achieving the maximum open space by covering the ground truth paths. The third part constrains the paths to the traversable areas.

\section{EXPERIMENTS}
In this section, we introduce the details of the implementation, describe the results, and compare MTG with other approaches. For the real-world experiment we combine MTG and a low-level motion planner AdaptiveON~\cite{liang2022adaptiveon} for global navigation as shown in Figure \ref{fig:all_trajectory}. Details of the real-world experiment can be found in Appendix~\cite{appendix} \ref{sec:realworld}. We also analyze the confidence of generated trajectories and the generalizability of our approach in Appendix~\cite{appendix} \ref{sec:analysis}.

\subsection{Implementation}
During experimentation, we deployed the model on different robots, a Clearpath Husky and a Boston Dynamics Spot. The major perceptive sensor is a 16-channel Lidar, Velodyne VLP-16, with a 3Hz frequency. we choose $\N_l = 3$ frames in the experiment, based on the prior research~\cite{liang2021crowd, densecavoid} that 3 previous frames are enough to encode the information of dynamic obstacles for collision avoidance. To keep the observation at the same time, we choose $N_v=10$ for velocities based on a 10Hz frequency measured by the robots' odometers. Our training and testing datasets are collected in a campus environment, as shown in  Appendix~\ref{sec:app_dataset}; the two datasets are collected from very different areas.  The training dataset contains three parts: 1) The original observation data, including Lidar and velocities. 2) Based on the perception, we build a traversability map, which is only used as the ground truth to guide the training. We briefly described the generation of the traversability map in Appendix~\ref{sec:traversable_map}. 3) Based on the map, we sample multiple diverse targets and apply an A* planner to generate raw ground truth trajectories. The targets are sampled 15m away, considering the robot's speed is 1m/s and the trajectories are generated for the next 15 seconds. Details are in Appendix~\cite{appendix} \ref{sec:app_trajectory}

As Figure \ref{fig:pipeline} shows, the encoder encodes perception information to a hidden vector $\c$, which is also the condition value of the decoder. It contains two sub-models. One is the Lidar model which composes multiple 3D convolution layers to process the stacked Lidar data. In this work, we use PointCNN~\cite{hua2018pointwise} to process the Lidar data with a 0.08m voxelization radius. The other model is a velocity model, including three consecutive fully connected layers, to process the velocity data. Then these two encoded perception data are concatenated and processed by several fully connected layers to generate the hidden vector $\c$. The $A_i(\c)$ and $b_i(\c)$ are Linear layers with the input, $\c$. The  $g(\cz)$, is processed by an attention model where the inputs are the processed Gaussian distribution vectors and the outputs are $\c_s$. Concatenated with the encoded condition $\c$, we have $\c_i = \set{\c_s, \c}$ as the input of the condition to the decoder. The decoder is composed of a sequence of GRU cells and outputs a sequence of $\set{\Delta x_n, \Delta y_n}$, which are accumulated to waypoints $p_n = (x_n, y_n)$ with the initial position of $\set{0, 0}$. The details of the architecture can be found in Appendix~\cite{appendix} \ref{sec:architecture_details}. In each generated trajectory, there are 16 waypoints, based on the furthest distance of 15m. The training is processed by an NVIDIA RTX A5000 GPU and an Intel Xeon(R) W-2255 CPU, and we use this machine for evaluation. The qualitative evaluation results between the ground truth A* and generated trajectories are in Appendix~\cite{appendix} \ref{sec:a_start_generated}.

\subsection{Comparing Results}
In this section, we qualitatively and quantitatively evaluate the performance of our method with other approaches.

\subsubsection{Quantitative Results}
The evaluation metrics include the following:

\textbf{Non-traversable rate:} Ratio of the generated trajectories lying on non-traversable areas. This metric is calculated by 
    \begin{align}
        r_n = \frac{1}{K} \sum_{k=1}^K \frac{g(m_\theta(o)_k, \Tilde{\ca})}{\abs{\tau_k}},
        \label{eq:traversable}
    \end{align}
    where $g(\cdot)$ is defined in Equation \ref{eq:definition} as the segments of trajectory $\tau_k$ lying on non-traversable areas.

\textbf{Coverage rate:} This metric measures the coverage of the traversable areas in the robot's perception by generated trajectories. In this project, we set the robot's perception to a 120-degree field of view in front of the robot. The assumption is that the ground truth trajectories covered all the traversable areas, which is true of most of the cases based on the traversable map we built. We measure the smallest Hausdorff distance between ground truth trajectories and the generated trajectories and use the following equation as the coverage metric:
\begin{align}
    r_c = \frac{1}{G} \sum_{i=1}^G \exp{-\min d_h(\tau_i, \widehat{\cy})},
    \label{eq:coverage_rate}
\end{align}
where $G$ is the number of all ground truth trajectories in the map and $\widehat{\cy}$ is the set of generated trajectories. $\tau_i$ is the $i_{th}$ of the ground truth trajectory.

\textbf{Diversity rate:} Diversity of the generated trajectories. This metric measures the Euclidean distance among the generated trajectories. This metric is a complement to the coverage metric.
\begin{align}
    r_d = \frac{1}{N^2} \sum_{i=1}^{N}\sum_{j=1}^{N} d_h(\hat{\tau_i}, \hat{\tau_j}), \;\; i\neq j.
\end{align}

\begin{table}[h!]
    \centering
    \begin{tabular}{ c c c c c } 
     \toprule
     \multirow{2}{*}{Methods} & \multirow{2}{*}{Non-traversable} & \multirow{2}{*}{Coverage} & Diversity  & Running \\
      &  &  &  ($10^{9}$) & Time (ms) \\
     \midrule
     Ground Truth & 0 & 1 & 28 & N/A \\
     CVAE~\cite{cvae}          & 0.14 & 0.63 & 7.0 & 5 \\
     DLOW~\cite{yuan2020dlow}          & 0.15 & 0.68 & \textbf{79.0} & \textbf{3} \\

     MTG$^1$(Ours) & 0.017 & 0.70 & 10.0 & 4 \\
     MTG(Ours)     & \textbf{0.013} & \textbf{0.72} & 14.0 & 5 \\
     \bottomrule
    \end{tabular}
    \caption{Comparison of CVAE~\cite{cvae}, DLOW~\cite{yuan2020dlow}, vanilla MTG, and complete MTG. The ground truth values are also provided as a reference. Our method achieves $6\%$ improvement in coverage of traversable areas using Equation \ref{eq:coverage_rate} and improves traversability of the trajectories using Equation ~\ref{eq:traversable}.}
    \label{tab:comparison}
    \vspace{-1em}
\end{table}

\begin{figure*}[ht]
     \centering
     \begin{subfigure}[b]{0.3\textwidth}
         \centering
         \includegraphics[width=\textwidth,height=0.8\textwidth]{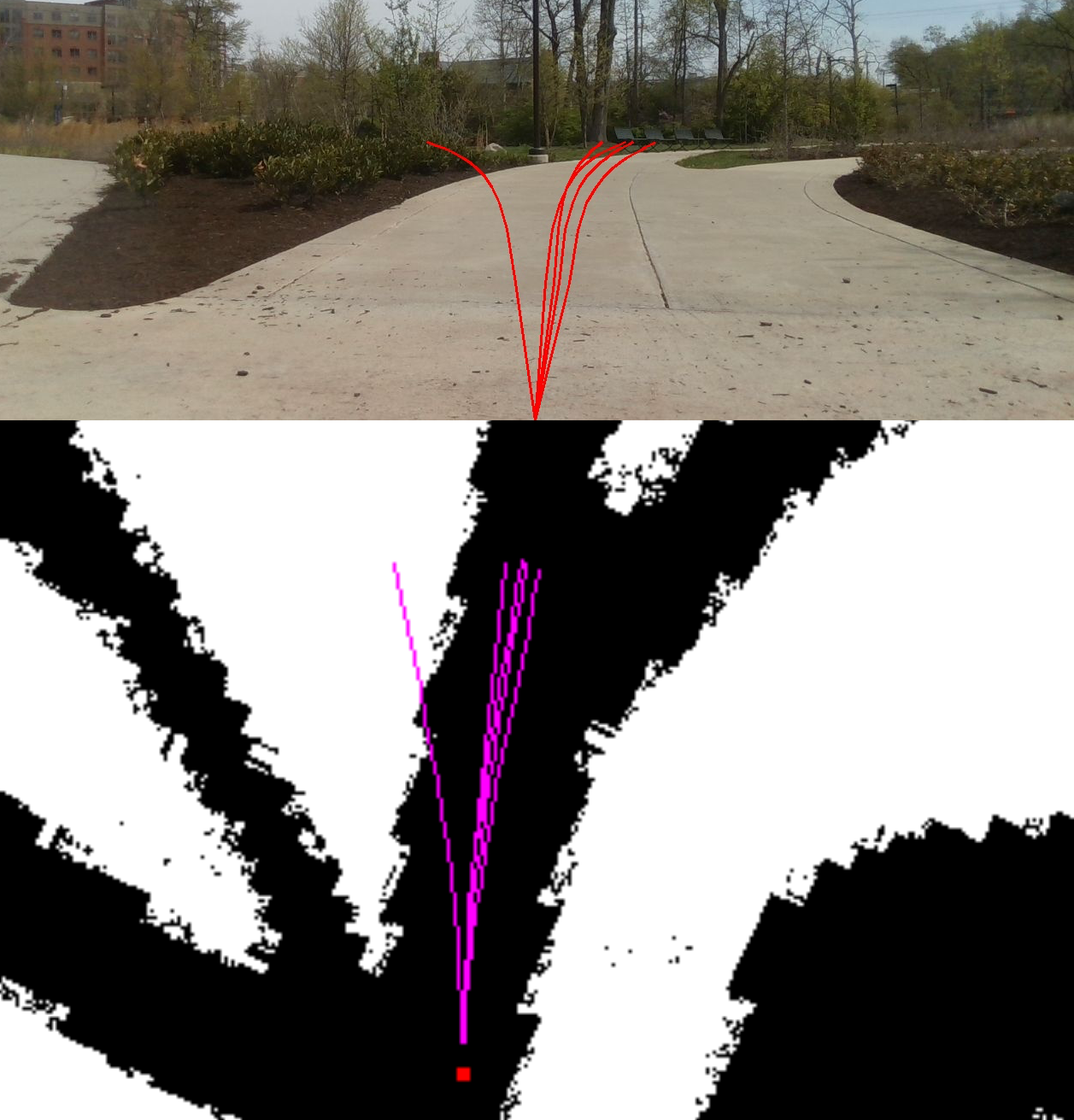}
         \caption{CVAE}
         \label{fig:cvae_grass}
     \end{subfigure}
     \hfill
     \begin{subfigure}[b]{0.3\textwidth}
         \centering
         \includegraphics[width=\textwidth,height=0.8\textwidth]{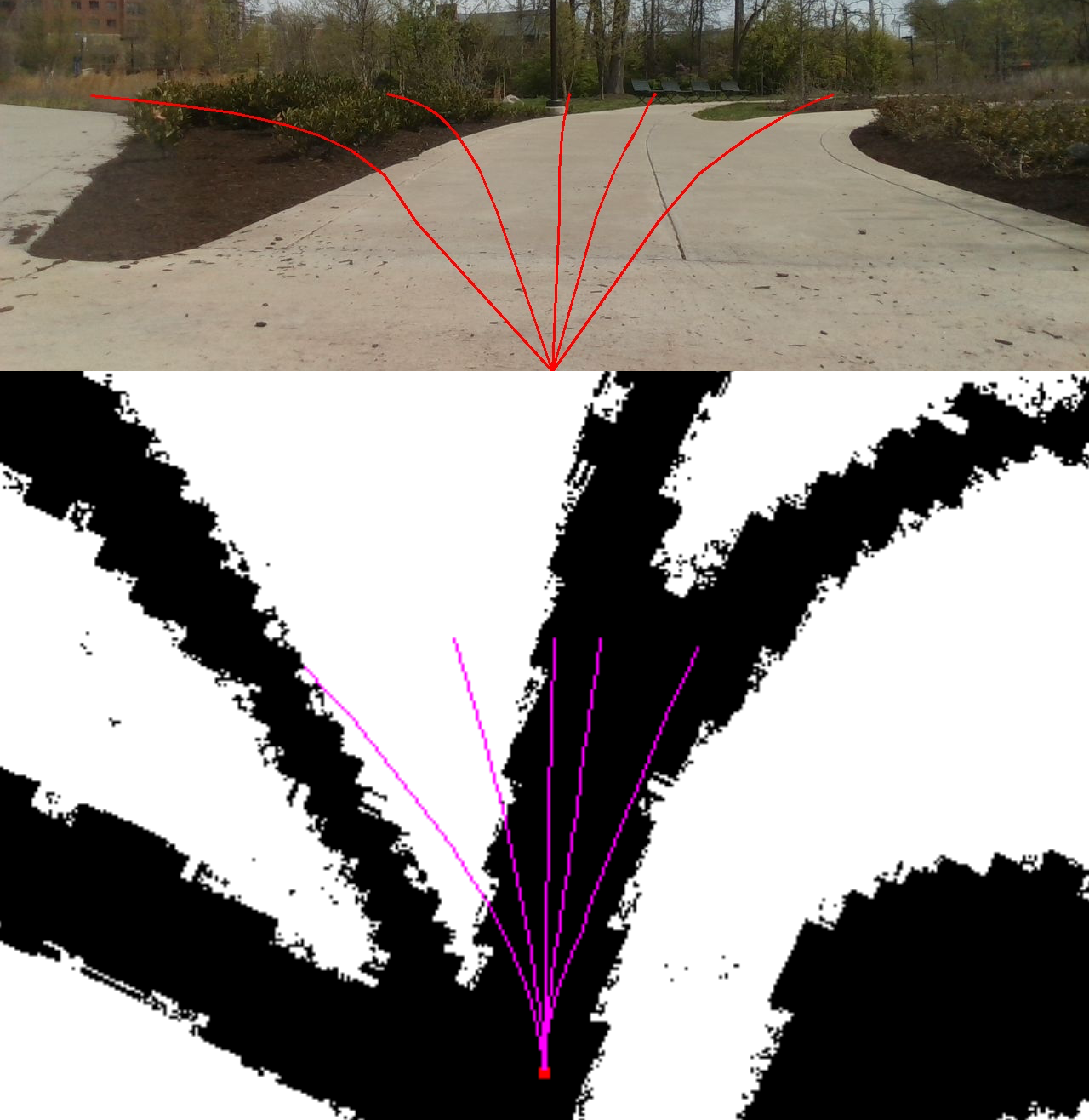}
         \caption{DLOW}
         \label{fig:dlow_grass}
     \end{subfigure}
     \hfill
     \begin{subfigure}[b]{0.3\textwidth}
         \centering
         \includegraphics[width=\textwidth,height=0.8\textwidth]{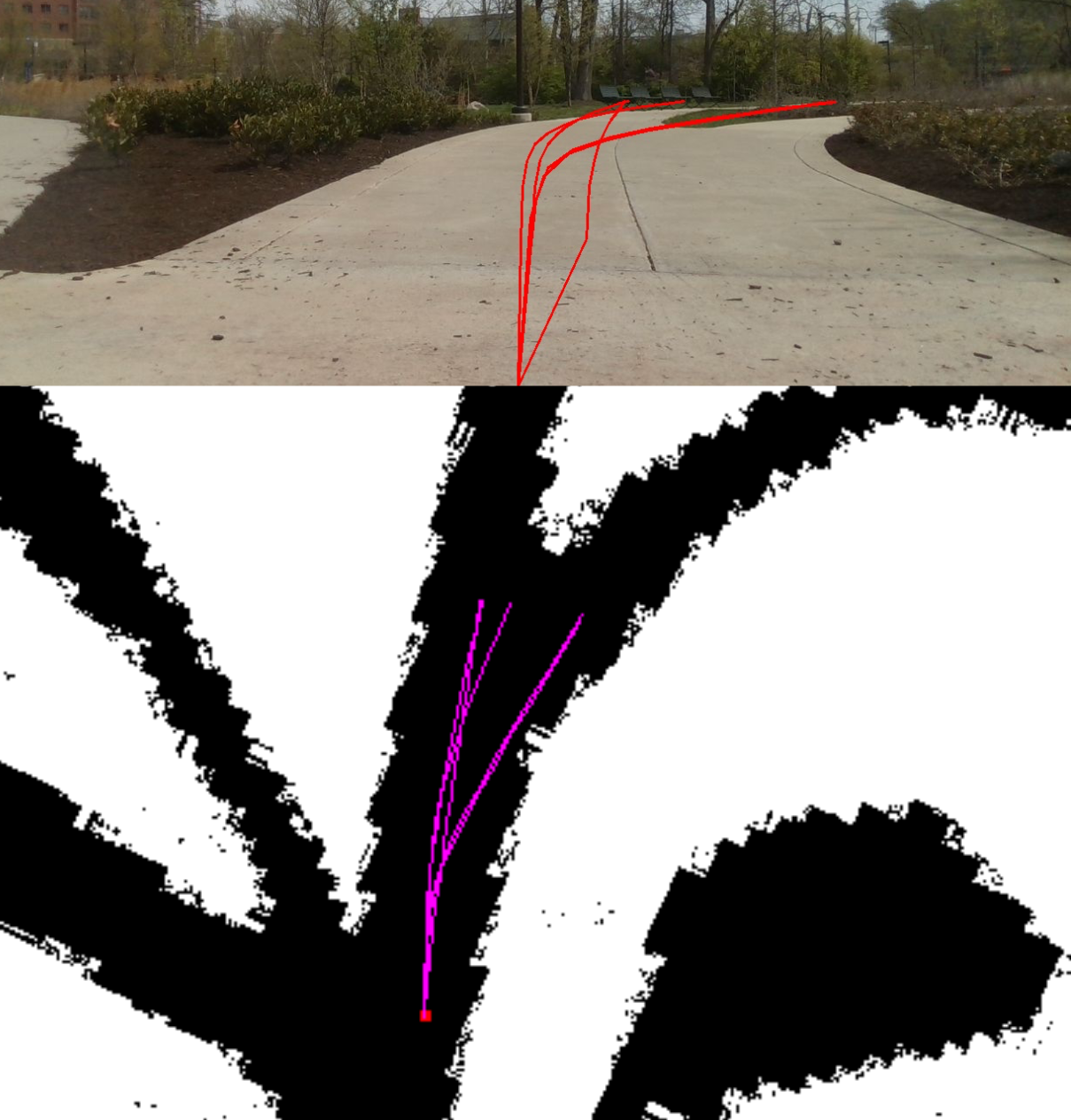}
         \caption{MTG}
         \label{fig:cot_grass}
     \end{subfigure}
    \caption{\textbf{The quality results} of CVAE~\cite{cvae}, DLOW~\cite{yuan2020dlow}, and MTG. The top row is from the robot’s view, and the bottom row is the birds-eye-view. White areas are non-traversable areas. The CVAE generates trajectories very similar to each other, while DLOW has a large diversity but is not good in terms of traversability. Our approach generates trajectories that cover mostly all traversable directions and lie on only traversable areas.}
    \label{fig:quality}
    \vspace{-1.5em}
\end{figure*}

\textbf{Running time:} $t$ in each step and the unit is seconds. 

We compared our method with ground truth A* paths, vanilla CVAE method~\cite{cvae}, and DLOW~\cite{yuan2020dlow}. As shown in Table \ref{tab:comparison}, CVAE doesn't have a diversity function or coverage constraints, so the output centers on very similar trajectories. Therefore, the coverage and diversity of CVAE are very low. The DLOW method evenly implies the diversity loss to separate the generated trajectories, so the diversity value is high. However, neither CVAE nor the DLOW method provides hard constraints on the non-traversable areas, leading to trajectories with large segments lying on non-traversable areas. MTG$^1$ is our method without global information of other trajectories. With the traversability loss, the generated trajectories mostly lie in the traversable areas. Because our diversity loss only applies to the trajectories nearest to the most effective trajectories (closest to the ground truth trajectories) and drives redundant trajectories close to effective trajectories, this model has better performance in coverage but lower diversity than DLOW. The final model, MTG, implies the information of other trajectories in each decoder; with the references from other trajectories, the model is easier to train with better coverage.

\subsubsection{Qualitative Results}
In this section, we qualitatively show the generation performance of our approach and compare trajectory quality with the other approaches.

\begin{figure}[ht]
    \centering
    \vspace{-0.25em}
    \includegraphics[width=\linewidth, height=0.5\linewidth]{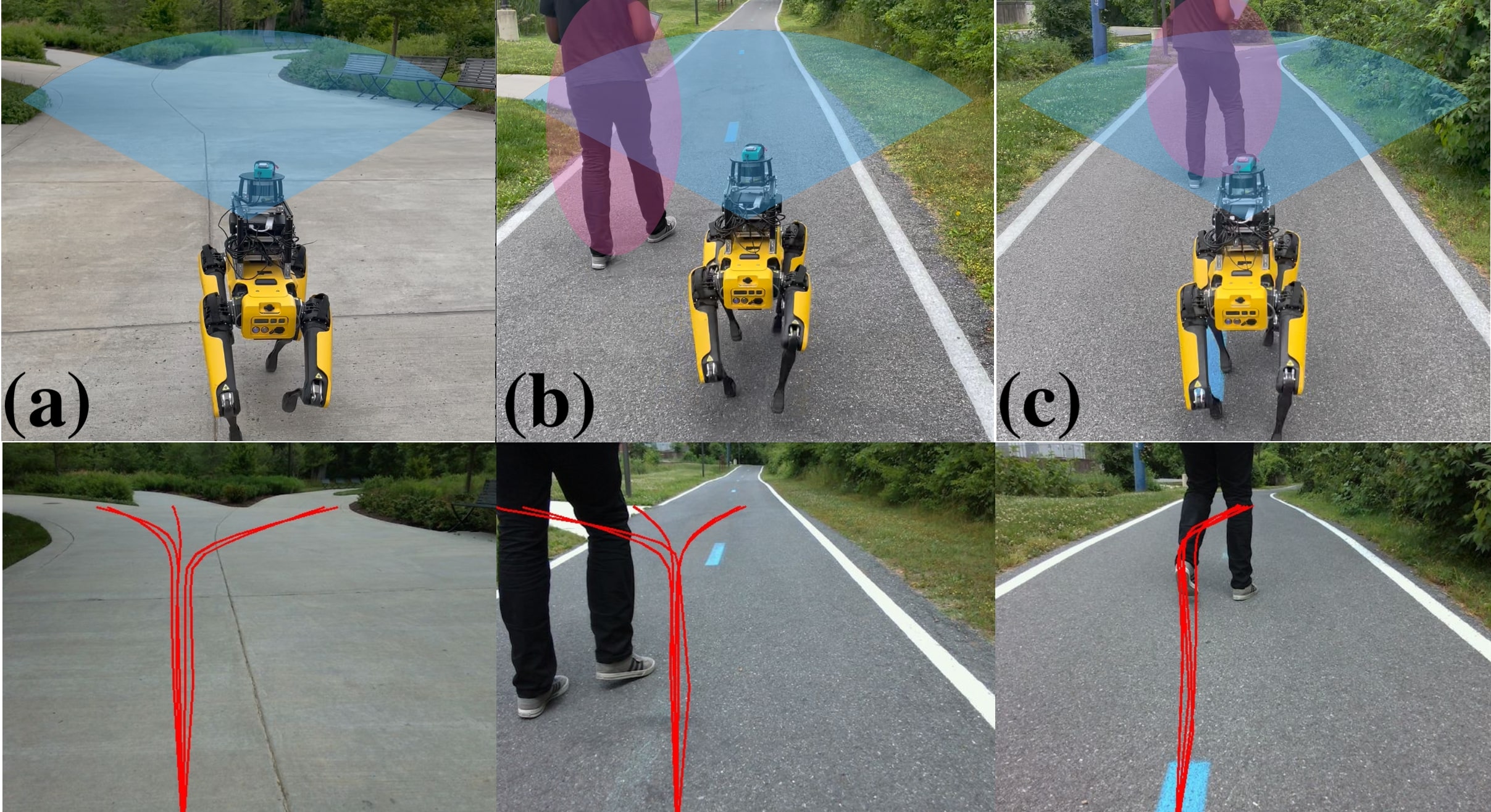}
    \vspace{-0.25em}
    \caption{Trajectory generation with different traversable areas; the blue area corresponds to the region visible to the robot, and the purple areas denote people.
    }
    \label{fig:spot}
    \vspace{-0.5em}
\end{figure}

As shown in Figure~\ref{fig:spot}, the bottom rows of the figures are from the robot’s view, and the top rows of the figures are captured from behind the robot. The generated trajectories are the red curves on the bottom row. Figure~\ref{fig:spot} (a) to (c) shows that, with different perception occlusions, the trajectory generator is still able to cover all diverse directions the robot can achieve on traversable areas. Figure~\ref{fig:spot} (a) shows the generated trajectories covering diverse future directions the robot can achieve in traversable areas. Figure~\ref{fig:spot} (b) shows the trajectory generation with partially occluded perception and Figure~\ref{fig:spot} (c) shows the trajectory generation when the only path in front of the robot is blocked. We can observe that the trajectories cover the traversable areas well and can also account for the influence of the dynamic obstacles. 

The trajectories generated by different methods are compared in Figure \ref{fig:quality}. The top row is the robot's view from its camera. The bottom row is the local map, where the white areas indicate the non-traversable regions, which are manually segmented in the local maps for evaluation. The purple and red curves are generated trajectories, starting from the bottom center of each image.

As shown in Figure \ref{fig:cvae_grass}, the trajectories generated by CVAE are mostly similar, and some of the trajectories lie on non-traversable areas. The DLOW method can generate very diverse trajectories, but, similar to CVAE and as shown in Figure \ref{fig:dlow_grass}, the trajectories cannot avoid non-traversable areas well. Figure \ref{fig:cot_grass} shows the trajectories from the MTG method, which performs better in covering the traversable areas and not infringing on the non-traversable zones.

\section{Conclusion and Future Work}
We present MTG, a novel mapless trajectory generation method for autonomous mobile robots. We introduce an innovative trajectory generation architecture that incorporates a Conditional Variational Autoencoder (CVAE) with an attention mechanism. We also propose novel loss functions that account for traversability, diversity, and coverage constraints. We demonstrate superior performance in terms of coverage of traversable areas and feasibility of the trajectories compared to SOTA methods. This work can be used in the future for extensive global planning tasks. 

This work has some limitations. The current dataset only provides a front view of the Lidar, thereby limiting the robot's ability to generate trajectories behind it. The traversability map also requires more manual labor before training. We plan to address these issues in future work.

\section*{Acknowledgement}
\vspace*{-0.5em}
This research was supported by Army Cooperative Agreement W911NF2120076 and ARO grants  W911NF2310046 and W911NF2310352.
\vspace*{-0.5em}

\bibliographystyle{IEEEtran}
\bibliography{ref}

\newpage
\clearpage

\section{Appendix}

\subsection{Real-world Experiment}
\label{sec:realworld}
For testing, a real-world experiment is done by combining MTG with a low-level AdaptiveON~\cite{liang2022adaptiveon} motion planner for global navigation, as in Figure \ref{fig:all_trajectory}. We choose the generated trajectory with the last waypoint closest to the target as the following trajectory for the local planner. In each step, the robot takes the nearest next waypoint in the trajectory as the next goal. The model runs in the timestep around 0.01s on an onboard machine with an Intel i7 CPU and one Nvidia GTX 1080 GPU.

\subsection{Analysis}
\label{sec:analysis}
\begin{figure*}[ht]
\centering
\includegraphics[width=0.9\linewidth,height=0.5\linewidth]{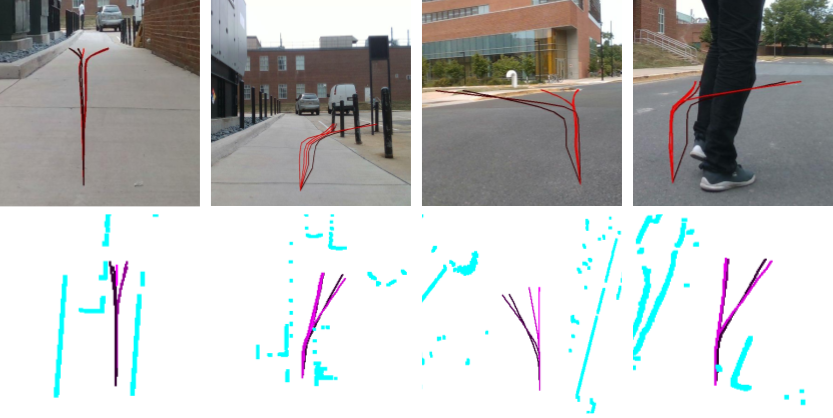}
\caption{Trajectory Confidence: The top row shows the camera view of the generated trajectories and the bottom row shows the bird-eye-view of the trajectories. The Cyan color represents the obstacles detected from the middle channel of the 3D Lidar.}
\label{fig:confidence}
\end{figure*}

\textbf{Confidence of Trajectories: } The trajectories are generated by the embedding Gaussian distributions, which are inputs of the decoder, and we train each distribution to cover one traversable direction so the standard deviation can tell the confidence of the trajectory. When the standard deviation is large, the confidence is low because the distribution is not certain of the mean value. In our formulation, the variances for each trajectory can be calculated as $\v_c = A \v A^T$, where $\v$ is the variance of the embedding. $\v_c = \set{\v_1,...,\v_k}$, where $\v_k$ and $A$ are defined in Section~\ref{sec:attention_approach}. The generated trajectories are shown in Figure \ref{fig:confidence}; the higher the variance, the lower the confidence and the darker the trajectory is. As the following figures show, when the trajectories are near the wall or other non-traversable areas, the confidence is low. In addition, the last row shows when the person is very close to the robot. Although the model can still ignore the person, the confidence of the trajectories (passing through the person) gets lower.

\begin{figure}[ht]
    \centering
    \includegraphics[width=\linewidth]{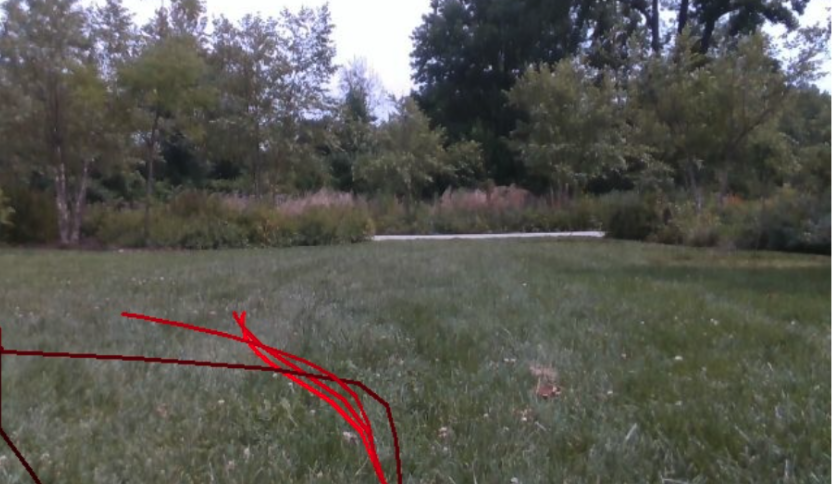}
    \includegraphics[width=\linewidth]{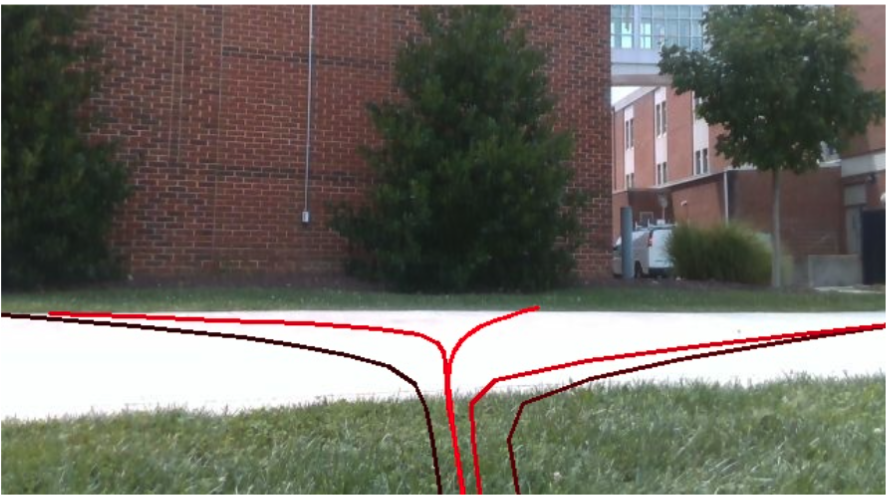}
    \caption{The out-of-distribution cases: The top figure shows the robot fully in a non-traversable area with no traversable area around it. The bottom figure shows the robot leaving the non-traversable area, where the robot can still generate good trajectories.}
    \label{fig:ood}
\end{figure}

\textbf{Generalization: } As shown in the above results, our model is well-generalizable to the campus environment. However, for completely different scenarios like cities or woods, we don’t expect our method to generalize very well to those out-of-distribution scenarios, which we believe is reasonable for most learning-based approaches. In addition to the out-of-distribution scenarios, because our model is not trained in non-traversable areas, the model cannot perform well in fully non-traversable areas. In Figure~\ref{fig:ood}, we demonstrate that when the robot operates within non-traversable zones, its trajectories exhibit significant randomness. However, as the robot exits these zones, its paths realign within the traversable areas.

\subsection{Traversability Map}
\label{sec:traversable_map}

Before building 2D maps, we use Lio-sam~\cite{shan2020lio} to build 3D maps by running robots on the campus. Then we remove the noisy points and calculate the surfaces of the area. Then large cliffs, bushes, and buildings are detected and marked as non-traversable areas. Next, we press the 3D maps into 2D maps, creating traversable maps with a resolution of 0.1m. Finally, we manually label some missing non-traversable areas by combining satellite maps.

\subsection{Trajectory Selection}
\label{sec:app_trajectory}

The trajectory length is empirically determined by our robot’s speed and the perceptive range of the Lidar sensor. On the one hand, we would like to generate long trajectories to provide the robot with proper guidance for future directions. On the other hand, unnecessarily long trajectories may lead to occluded areas or out of the perception of the Lidar sensor. Our two robots drive around 1-2m/s, and 10-20 meters are enough for robots to drive for 10-20 seconds before the next trajectory generation. Therefore, we chose 15 meters as our trajectory length.



\subsection{Qualitative Evaluation}
\label{sec:a_start_generated}

\begin{figure*}[ht]
\centering
\includegraphics[width=0.24\linewidth,height=0.25\linewidth]{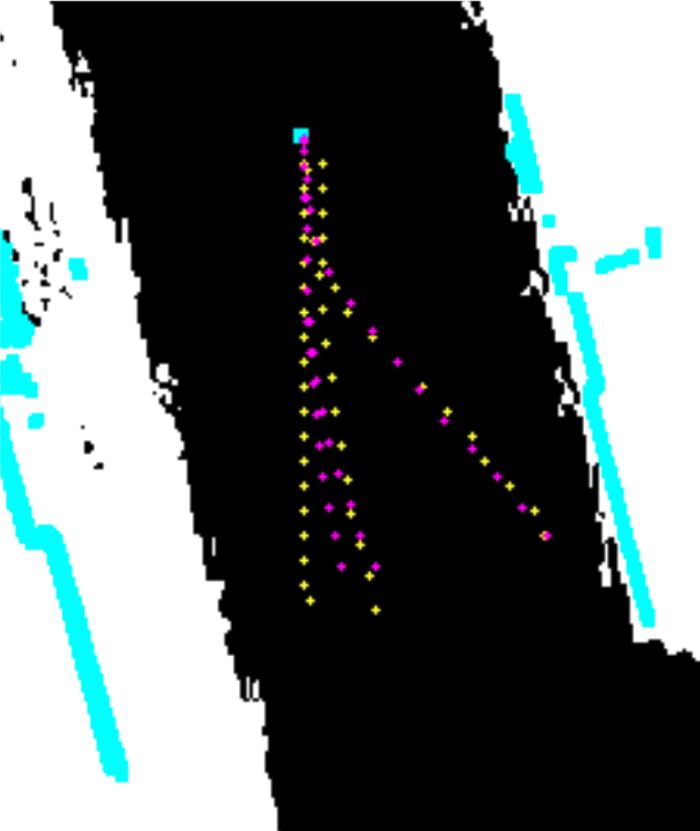}
\includegraphics[width=0.24\linewidth,height=0.25\linewidth]{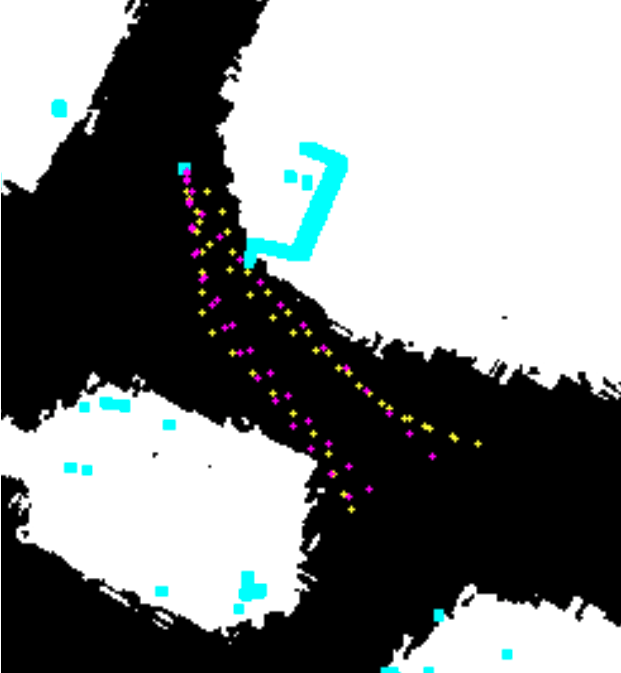}
\includegraphics[width=0.24\linewidth,height=0.25\linewidth]{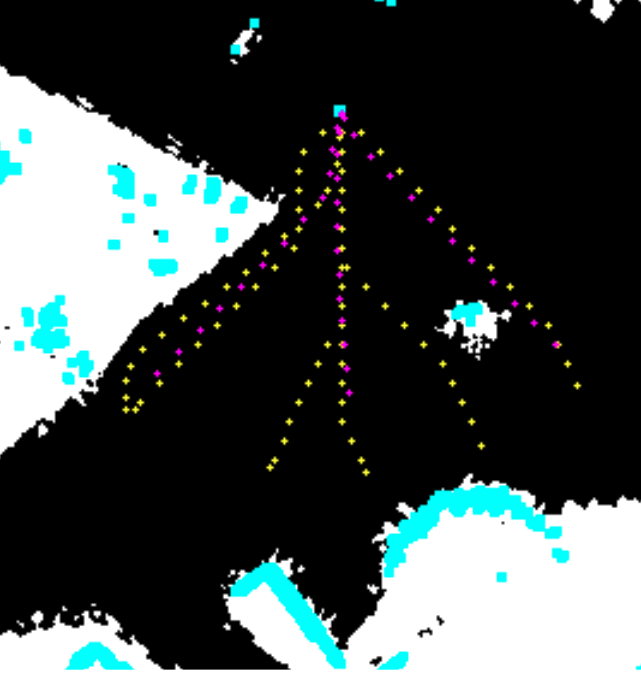}
\includegraphics[width=0.24\linewidth,height=0.25\linewidth]{figures/app/evaluate/evaluate_2.png}
\caption{Evaluation: A* ground truth paths are yellow trajectories and generated paths from MTG are purple trajectories. The white are non-traversable areas and cyan is the obstacle detected by the middle channel of the 16-channel Velodyne Lidar.}
    \label{fig:evaluation}
\end{figure*}

Figure \ref{fig:evaluation} shows the birds-eye-view of the generated trajectories (purple) and ground truth trajectories (yellow and generated by the A* algorithm). Cyan represents Lidar points from the middle channel (8th channel) of a 16-channel Lidar scan for reference. The white areas are all non-traversable areas. We can see the purple trajectories generated by our model are very smooth and cover the traversable areas (black areas), similar to the yellow A* paths.

\subsection{Architectural Details}
\label{sec:architecture_details}

\begin{figure*}[ht]
    \centering
    \includegraphics[width=0.8\linewidth]{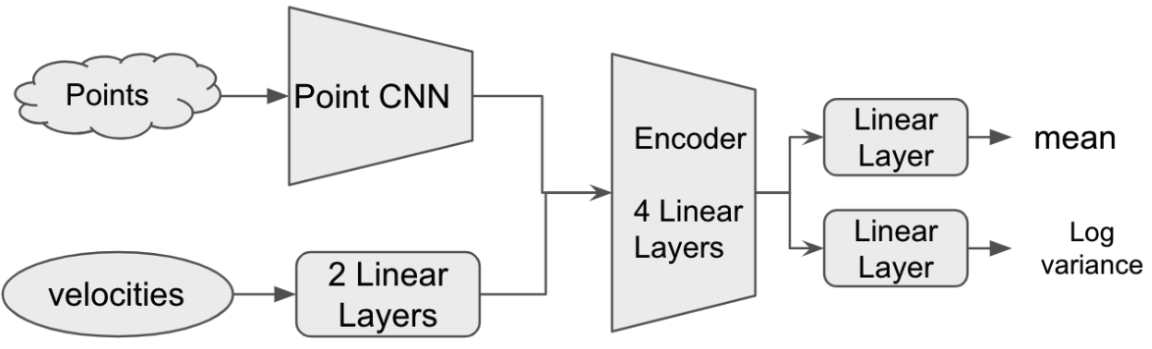}
    \caption{Perception Models}
    \label{fig:perception_model}
\end{figure*}

Perception models in Figure \ref{fig:perception_model}: The output of PointCNN has dimension 512, and it concatenates with the velocity embeddings, with dimension 256 as input of the Encoder. The output of the encoder has dimension 512. The function $h_\psi(\z)$ contains two linear layers, which output dimension 256. The $A(\cdot)$ and $b(\cdot)$ functions are all single linear layers, and the outputs have the dimension of (512 x trajectory number).

\begin{figure*}[ht]
    \centering
    \includegraphics[width=0.8\linewidth]{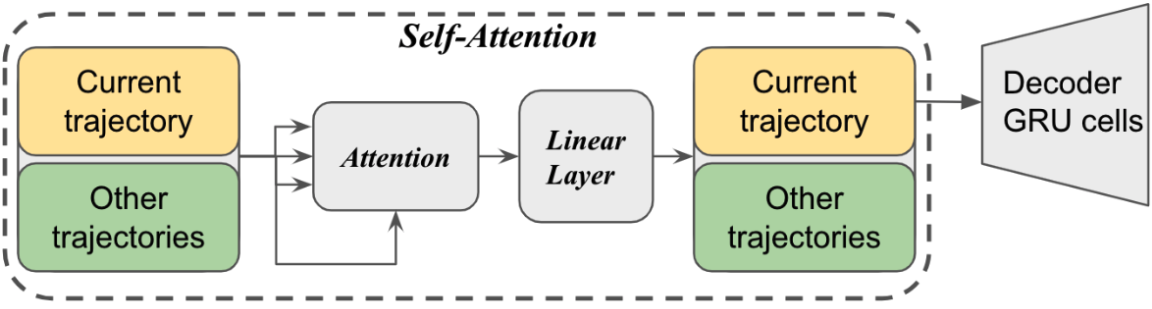}
    \caption{Self-Attention Model}
    \label{fig:attention_models}
\end{figure*}

Self-Attention Model: As in Figure \ref{fig:attention_models}, the trajectory embedding keeps the dimension 512.

\subsection{Dataset}
\label{sec:app_dataset}
\begin{figure*}
    \centering
    \includegraphics[width=\linewidth,height=0.7\linewidth]{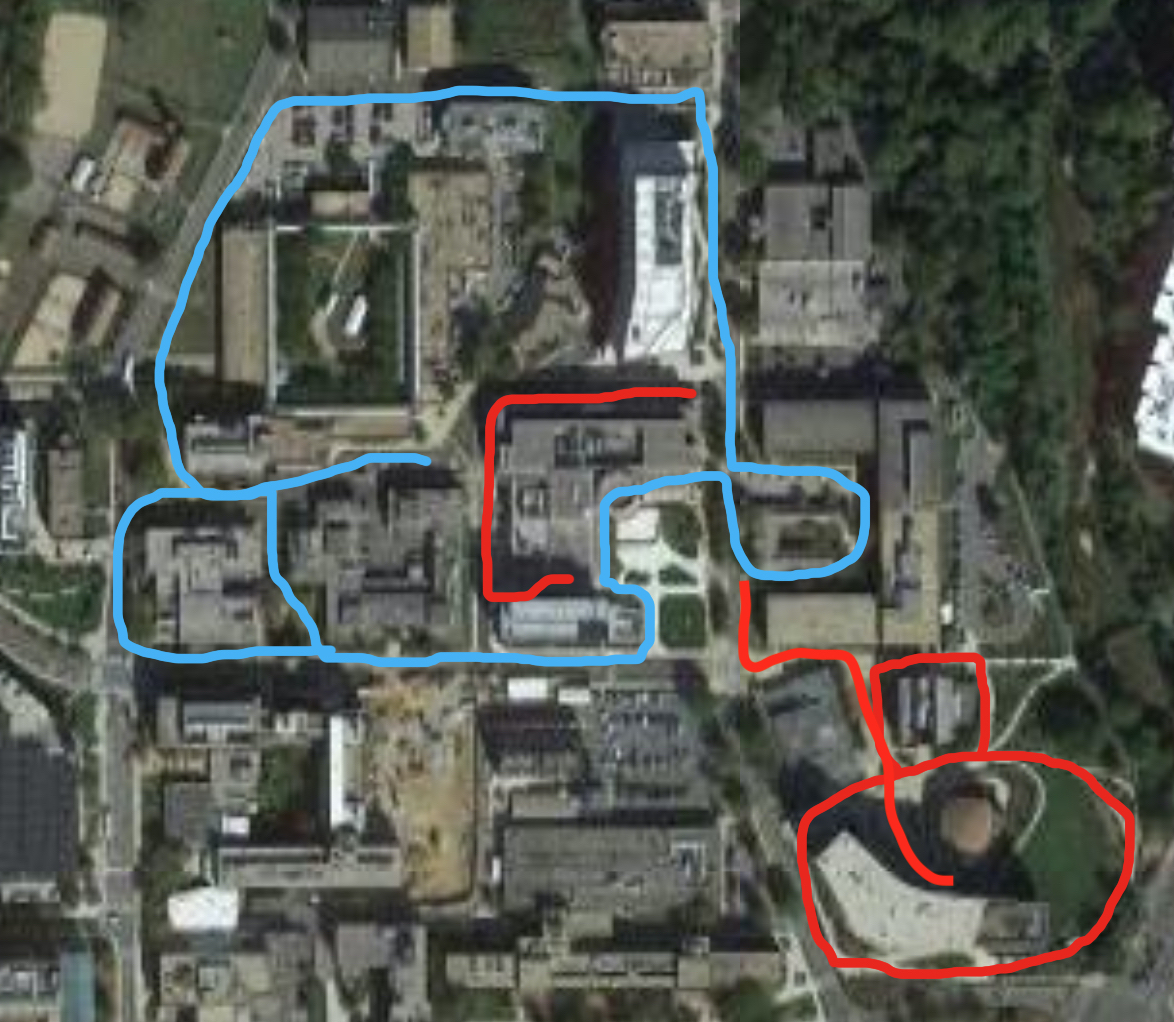}
    \caption{The data is collected at a university campus. The blue trajectories are for training and the red are for testing.}
    \label{fig:satellite_map}
\end{figure*}

The dataset is collected on a university campus. As shown in the satellite map \ref{fig:satellite_map}, the blue trajectories are for the training dataset and the red trajectories are for the testing dataset. In total, there are 2076 frames with all required perception data and local maps.

\end{document}